\documentclass[journal]{IEEEtran}

\usepackage{cite}
\usepackage{amsmath,amssymb,amsfonts}
\usepackage{algorithm}
\usepackage{graphicx}
\usepackage{makecell}
\usepackage{multirow}
\usepackage{color}
\usepackage{algpseudocode}

\usepackage{textcomp}
\usepackage{wrapfig}
\usepackage[colorlinks,
            linkcolor=red,
            anchorcolor=blue,
            citecolor=green 
            ]{hyperref}

\begin{document}

\title{AngularFuse: A Closer Look at Angle-based Perception for Spatial-Sensitive Multi-Modality Image Fusion}

\author{Xiaopeng Liu, Yupei Lin, Sen Zhang, Xiao Wang, Yukai Shi, Liang Lin,~\IEEEmembership{Fellow, IEEE}
\thanks{ 

X. Liu and Y. Shi are with School of Information Engineering, Guangdong University of Technology, Guangzhou, 510006, China (email: xiaopeng22@foxmail.com; ykshi@gdut.edu.cn) . 

S. Zhang is with TikTok, ByteDance Inc, Sydney, NSW 2000, Australia (email: senzhang.thu10@gmail.com).

X. Wang is with School of Computer Science, Anhui University, Hefei, 230000, China (email: xiaowang@ahu.edu.cn).

Y. Lin and L. Lin are with School of Computer Science and Engineering, Sun Yat-sen University, Guangzhou, 510006, China (email: yupeilin2388@gmail.com; linliang@ieee.org).
}

}
\maketitle

\begin{abstract}
Visible–infrared image fusion is crucial in key applications such as autonomous driving and nighttime surveillance. Its main goal is to integrate multimodal information to produce enhanced images that are better suited for downstream tasks. Although deep learning based fusion methods have made significant progress, mainstream unsupervised approaches still face serious challenges in practical applications. Existing methods mostly rely on manually designed loss functions to guide the fusion process. However, these loss functions have obvious limitations. On one hand, the reference images constructed by existing methods often lack details and have uneven brightness. On the other hand, the widely used gradient losses focus only on gradient magnitude. To address these challenges, this paper proposes an angle-based perception framework for spatial-sensitive image fusion (AngularFuse). At first, we design a cross-modal complementary mask module to force the network to learn complementary information between modalities. Then, a fine-grained reference image synthesis strategy is introduced. By combining Laplacian edge enhancement with adaptive histogram equalization, reference images with richer details and more balanced brightness are generated. Last but not least, we introduce an angle-aware loss, which for the first time constrains both gradient magnitude and direction simultaneously in the gradient domain. AngularFuse ensures that the fused images preserve both texture intensity and correct edge orientation. Comprehensive experiments on the MSRS, RoadScene, and M3FD public datasets show that AngularFuse outperforms existing mainstream methods with clear margin. Visual comparisons further confirm that our method produces sharper and more detailed results in challenging scenes, demonstrating superior fusion capability.
\end{abstract}


\section{Introduction}
\IEEEPARstart  {I}mage fusion is a technique that integrates information from multiple image sources. It can generate composite images with greater visual expressiveness and functional value. The key point is to utilize the complementary characteristics of images from different modalities to compensate for the deficiencies of a single imaging mode~\cite{intr1,intr2,intr3}. This technique has demonstrated significant value in several key areas such as autonomous driving, medical diagnosis, and remote sensing monitoring.

In recent years, deep learning has become the mainstream method in the field of image fusion. The key issue of visible-infrared image fusion (VIF) is the lack of real reference images. Therefore, unsupervised learning has become the main paradigm in visible-infrared image fusion research. Researchers primarily focus on designing loss functions to ensure the consistency between the fused image and the source images.

In unsupervised VIF methods, the design of the loss function is crucial. The loss typically includes pixel-level intensity loss (e.g.,$\mathcal L_{int})$ and gradient loss (e.g., $\mathcal L_{grad}$). However, current mainstream methods have many limitations in the design of pixel-level intensity loss and gradient loss. For example:
\begin{itemize}
\item[$\bullet$] \emph{Linear weighted loss}~\cite{li2021rfn} integrates the pixel differences between the fused image and the weighted source images by constraining with the L2 norm, defined as:
\begin{figure*}[!t]
\centering
\includegraphics[width=0.99\linewidth]{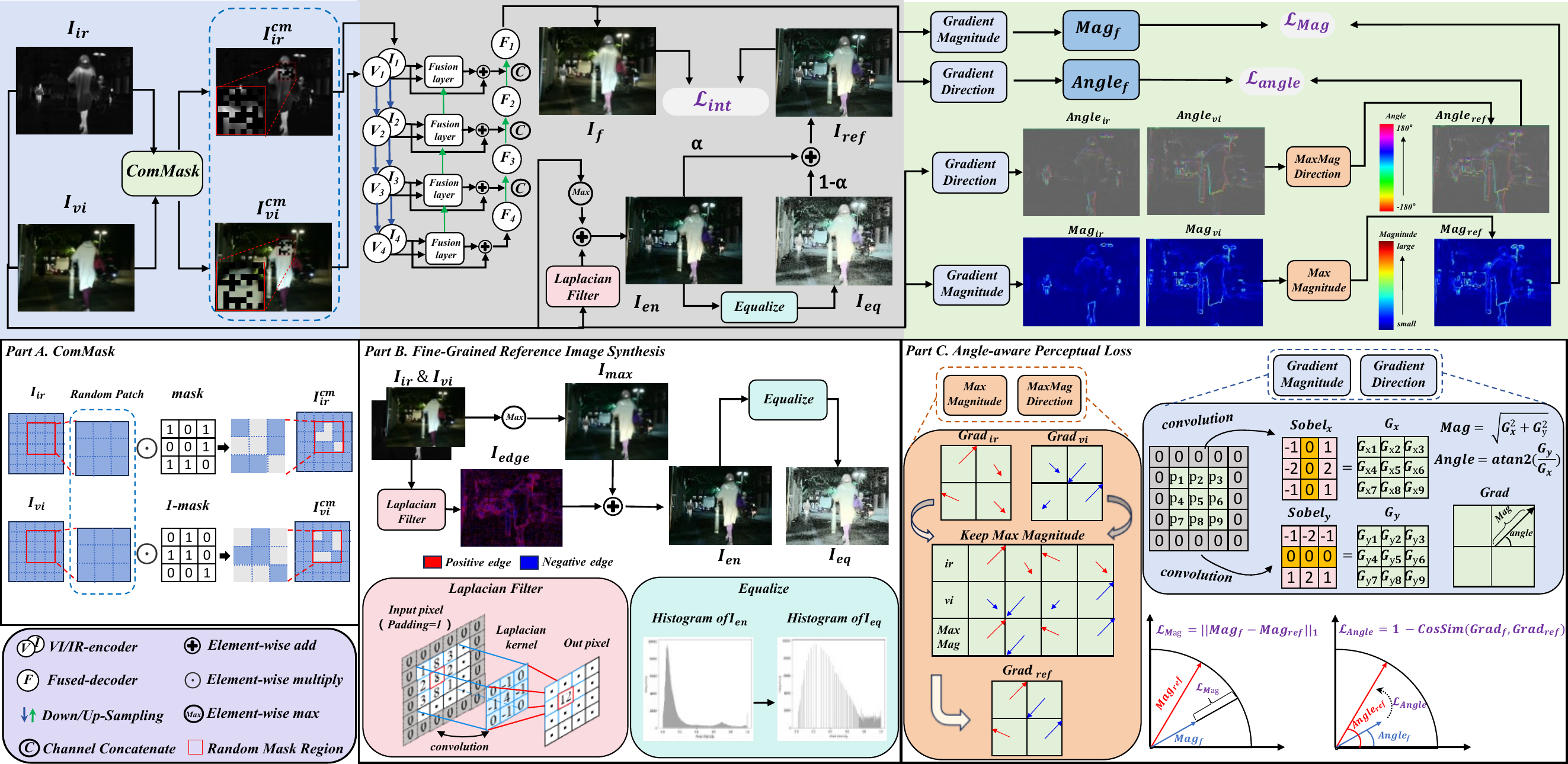}
\caption{We call for a closer look at angle-based perception for infrared- and visible- image fusion. The framework contains three parts. (a) Complementary Mask Generation: It takes $I_{ir}$ and $I_{vi}$ as input and produces incomplete views ${I}^{{cm}}_{{ir}}$ and ${I}^{{cm}}_{{vi}}$ to force the model to borrow information from the other modality. These two views are then sent to the fusion network to obtain $I_f$. (b) Fine-Grained Reference Image Synthesis: To obtain a reliable reference, we first apply a Laplacian filter to $I_{ir}$ and $I_{vi}$ to extract edge detail $I_{edge}$. We then blend $I_{edge}$ with the high-intensity image $I_{max}$ to produce $I_{en}$. Histogram equalization is applied to obtain $I_{eq}$ and mitigate uneven brightness. The reference image $I_{ref}$ is generated by weighted fusion of $I_{eq}$ and $I_{en}$ with ratio $\alpha$ and is used to compute the reference loss $\mathcal{L}_{\text{int}}$. (c) Angle-aware Perception: This module computes gradient magnitudes for infrared and visible images and builds a reference gradient by element-wise maximum. It then compares this reference with the gradient of the fused image $I_f$ and defines two loss terms $\mathcal{L}_{\text{Mag}}$ and $\mathcal{L}_{\text{Angle}}$ to improve structural fidelity and detail sharpness. }
\label{fig1}
\vspace{-3mm}
\end{figure*}
\begin{equation}
\mathcal{L}_{\mathrm{int\_linear}} = \left\| I_{f} - (w_1 I_{ir} + w_2 I_{vi}) \right\|_2
\end{equation}
where ${I}_{{f}}$ denotes the fused image, $I_{{ir}}$ and $I_{{vi}}$  represent the infrared and visible images, respectively. $w_1$ and $w_2$ are the corresponding weights. However, this linear weighted design does not consider the characteristics of different source image modalities. It is difficult to adapt to the feature distributions of various scenes and is prone to losing details in the fused image.
\item[$\bullet$] \emph{Modal Prior Based Loss}~\cite{fsgan} takes into account the differences between the two modalities. The pixel-level intensity loss of this method constrains the pixel differences between the fused image and the infrared image using the L2 norm:
\begin{equation}
\begin{aligned}
\mathcal{L}_{\mathrm{int\_mp}} = \left\| I_{f} - I_{ir} \right\|_2 \\
\mathcal{L}_{\mathrm{grad\_mp}} = \xi \left\| \nabla I_{f} - \nabla I_{vi} \right\|_2
\end{aligned}
\end{equation}
where \(I_f\) denotes the fused image, \(I_{\text{ir}}\) and \(I_{\text{vi}}\) represent the infrared and visible images, respectively. \(\nabla\) represents the gradient extraction operation, and \(\xi\) is the loss ratio adjustment coefficient. However, this method still has limitations. It merely assumes that the visible image carries detail features while the infrared image contains only intensity information. 
\item[$\bullet$] \emph{Multi-Modal Loss}~\cite{ganmcc}  designs a more comprehensive loss function. It considers both pixel-level intensity and gradient losses for the two modalities:
\begin{equation}
\begin{aligned}
\mathcal{L}_{\mathrm{int\_mm}} = \beta_1 \left\| I_{f} - I_{\mathrm{ir}} \right\|_2 + \beta_2 \left\| I_{f} - I_{\mathrm{vi}} \right\|_2  \\
\mathcal{L}_{\mathrm{grad\_mm}} = \beta_3 \left\| \nabla I_{f} - \nabla I_{\mathrm{ir}} \right\|_2 + \beta_4 \left\| \nabla I_{f} - \nabla I_{\mathrm{vi}} \right\|_2
\end{aligned}
\end{equation}
where \(I_f\) denotes the fused image, \(I_{\text{ir}}\) and \(I_{\text{vi}}\) represent the infrared and visible images, respectively. \(\nabla\) represents the gradient extraction operation, and \(\beta_i\) are the coefficients for adjusting the proportions of different losses. Under low-light conditions, the weight cannot adapt and the texture of the visible image will lost. Forcing reliance on its gradient information will introduce noise.

\item[$\bullet$] \emph{Maximal Preservation Strategy}\cite{Zhao_2023_CVPR, Zhao_2024_CVPR} adopts a maximal preservation strategy. They directly let the fused image retain the most prominent features of the source images at both the pixel and gradient levels:
\begin{equation}
\begin{aligned}
\mathcal{L}_{\mathrm{int\_max}} &= \left\| I_{f} - \max(I_{\mathrm{ir}}, I_{\mathrm{vi}}) \right\|_1 \\
\mathcal{L}_{\mathrm{grad\_max}} &= \left\| |\nabla I_{f}| - \max(|\nabla I_{\mathrm{ir}}|, |\nabla I_{\mathrm{vi}}|) \right\|_1
\end{aligned}
\end{equation}
where \(I_f\) denotes the fused image, \(I_{\text{ir}}\) and \(I_{\text{vi}}\) represent the infrared and visible images, respectively. \(\nabla\) represents the gradient extraction operation, and max(·) denotes the element-wise maximum operation. This strategy can effectively highlight crucial features. However, it ignores complementary information by retaining only the maximum values.
\end{itemize}
Thus, the goal of the pixel-level intensity loss is to learn the content distribution and pixel-level intensity features of the source images. The gradient loss (\(\mathcal{L}_{\text{grad}}\)) enhances the expression of detail features by constraining the local differences between pixels. However, existing methods have two key limitations. First, the reference image constructed in the pixel-level intensity loss fails to fully represent the features of multi-modal fusion. Second, the gradient loss only focuses on the gradient magnitude information. This overlooks the directional property of the pixel-level gradient vector. To address these issues, the core contributions are reflected in the following three aspects: 
\begin{itemize}
\item[$\bullet$] We propose a complementary masking strategy for cross-modality. The model is guided to learn cross-modal feature completion and the network is enhanced in modeling the complementary relationships between modalities.
\item[$\bullet$] By combining Laplacian edge enhancement and histogram equalization, we dynamically construct a reference image with richer detail features. This approach significantly optimizes the supervision of the pixel-level intensity loss.
\item[$\bullet$] We incorporate gradient spatial consistency into optimization functions, suppressing the limitations of single-magnitude constraints. By designing an angle-aware perceptual loss that considers both magnitude and angle, we achieve optimal performance of the fused image in terms of fusion quality and texture structure.
\end{itemize}
\section{Related work}
\subsection{Infrared- and Visible- Image Fusion}
In image fusion, the improvement of neural network structures bring a great deal of progress. Researchers build deeper architecture~\cite{wl1,wl2,wl3,wl4,wl5,wl6} to extract finer details. Some methods introduce attention mechanisms~\cite{zyl1,zyl2,zyl3,zyl4,zyl5}, enabling the model to focus more precisely on important feature regions in the image. In the discriminator design, the use of a multi-scale discriminator allows for the assessment of the realism of the fused image at different scales~\cite{gan0,gan1,gan2,gan3,gan4}. In addition, some studies combine Transformer with CNN~\cite{jh1,jh2,jh3,jh4,jh5,jh6,jh7}. The self-attention mechanism of Transformer captures long-range dependencies, while CNN refines local features.

More recently, the combination of the visible-infrared image fusion (VIF)~\cite{wu2025fully,li2025mulfs,xu2025hifusion} and image enhancement~\cite{chen2025faithdiff,yang2023reference,shi2022criteria,weng2025dronesr} further enhances the visibility. CROSE~\cite{XIAN2024123470} first employs a low-light enhancement technology for the image fusion task. NiteDR~\cite{shi2024nitedr} develops an image enhancement framework tailored for rainy nighttime driving scenes. Especially, CrossFuse~\cite{shi2025crossfuse} introduces a novel ranking strategy to receive effective representations from different datasets. DFVO~\cite{zhou2025dfvo} makes an attempt to disentangle infrared from visible light to perform a fine-grained image fusion and enhancement. These innovative multi-task designs provide new ideas for the development of image fusion technology.

\subsection{Unsupervised Image Fusion Loss Design}
The initial linear weighted loss constrains the pixel differences between the fused image and the weighted source images using the L2 norm~\cite{li2021rfn}. Yet it ignores the traits of each modality and fails across scenes. FusionGAN~\cite{fsgan} adds a gradient loss to enforce detail. This method only assumes that the visible image carries detailed features while the infrared image contains only intensity information. The subsequent GANMcC~\cite{ganmcc} method designed a more comprehensive loss function, considering both the pixel-level intensity loss and gradient loss of the two modalities. The method proposed constraints on the coefficients of different losses. However, under low-light conditions, this weight design fails to meet the demand. Subsequently, some methods (such as CDDfuse~\cite{Zhao_2023_CVPR} and EMMA~\cite{Zhao_2024_CVPR}) adopted a maximum-value-preserving strategy, directly retaining the most prominent features in the fusion image at both the pixel and gradient levels. This strategy can effectively highlight important features, but it ignores the authenticity and reliability of the fusion results by only preserving the maximum values.

\section{Methodology}

\subsection{Overview}
Fig.~\ref{fig1} shows the overall framework of the proposed method. The framework consists of three parts: Complementary Mask Generation, Fine-Grained Reference Image Synthesis, and Angle-aware Perceptions. The Complementary Mask Generation (ComMask) module is designed to enhance the network capability of complementary modeling across modalities through a random complementary masking strategy. The Fine-Grained Reference Image Synthesis (FRIS) module constructs more expressive reference images to improve the guidance effect of pixel-level intensity loss. The Angle-aware Perception module introduces gradient vector consistency into the optimization objective and overcomes the limitation of traditional gradient loss that only focuses on magnitude. This loss introduces directional constraints of gradient information to ensure that the fused image remains highly consistent with the source images in texture structure and edge orientation.
\begin{figure}[t]
\centering
\includegraphics[width=0.95\linewidth]{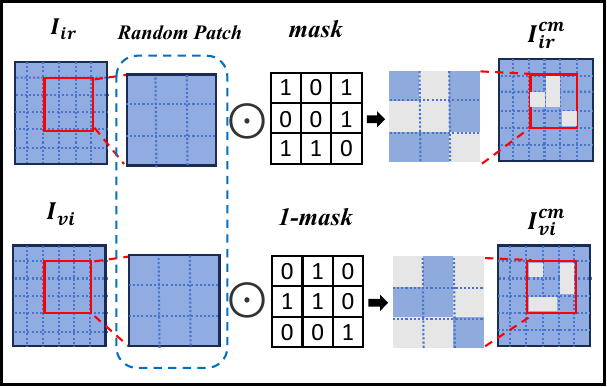}
\caption{The proposed Complementary Mask Generation (ComMask) module. First, a patch is randomly selected from the input infrared image $I_{ir}$ and the visible image $I_{vi}$. Then a complementary mask is applied to perform complementary occlusion, producing $I_{ir}^{cm}$ and $I_{vi}^{cm}$ with complementary spatial structures, for a cross-modality self-supervised learning.}
\label{fig1.2}
\vspace{-3mm}
\end{figure}
\subsection{Complementary Mask Generation}
The workflow of Complementary Mask Generation (ComMask) is shown in Fig.~\ref{fig1.2}. Specifically, for the input image pair (${I}_{ir}$, ${I}_{vi}$), a $k\times k$ square region $\mathcal{P}$ is randomly selected within the $H\times W$ spatial range. In the selected region  $\mathcal{P}$, random masking is performed. First, a random binary matrix ${R} \in \{0,1\}^{k\times k}$ is generated, where the proportions of 0 and 1 are each about 50\%. We set the infrared image mask as ${M}_{ir}(\mathcal{P}) = {R}$ and the visible image mask as ${M}_{vi}(\mathcal{P}) = {1} - {R}$. For the regions outside of $\mathcal{P}$, we keep the original information: ${M}_{ir}(\overline{\mathcal{P}}) = {M}_{vi}(\overline{\mathcal{P}}) = {1}$. Finally, we apply the masks to the input images:
\begin{equation}
{I}_{{ir}}^{{cm}}={I}_{{ir}}\odot{M}_{{ir}}, 
{I}_{{vi}}^{{cm}}={I}_{{vi}}\odot{M}_{{vi}}
\end{equation}
where $\odot$ denotes element-wise multiplication. The two masked inputs are partially incomplete, yet the missing regions are complementary across modalities. These masked infrared and visible images are then fed into the fusion network to yield the fused result ${I}_f$. 

Unlike traditional autoencoders~\cite{bachmann2022multimae}, our method focuses not only on single-modality reconstruction but also on the information interaction between the two modalities. This enables the model to effectively utilize complementary information to enhance fusion performance.
\begin{figure}[t]
\centering
\includegraphics[width=1\linewidth]{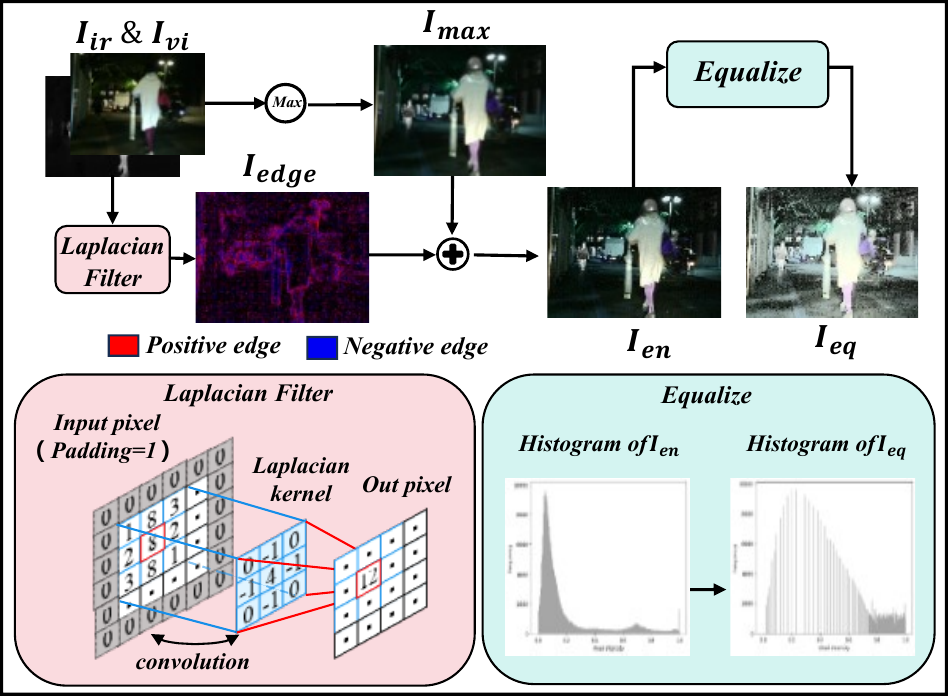}
\caption{The fine-grained reference image synthesis process. First, Laplacian convolution is applied to the infrared and visible images to extract the edge map $I_{edge}$. Then the high-intensity features $I_{max}$ are combined with $I_{edge}$ to obtain the enhanced image $I_{en}$ with rich textures. Finally, an equalization operation is applied to generate $I_{eq}$ with uniform brightness.}
\label{fig1.3}
\vspace{-3mm}
\end{figure}
\subsection{Fine-Grained Reference Image Synthesis}
As shown in Fig.~\ref{fig1.3}. We propose a more fine-grained reference image synthesis, whose construction process includes three key steps. First, we build a texture-enhanced image based on the Laplacian operator:
\begin{equation}
I_{\mathrm{en}} = \mathcal{T} \left( \nabla^2 (I_{\mathrm{ir}} + I_{\mathrm{vi}}) + \max(I_{\mathrm{ir}}, I_{\mathrm{vi}}) \right)
\end{equation}
where \( \nabla^{2} \) denotes the Laplacian operator and \( \mathcal{T}(\cdot) \) is a truncation function that keeps pixel values in the valid range [0, 255]. By injecting the extracted high-frequency component, the module sharpens texture expression. Then, we apply a histogram-equalization module to reduce brightness imbalance:
\begin{equation}
I_{eq} = \mathcal{H}( I_{en})
\end{equation} 
where $\mathcal{H}(\cdot)$ denotes the histogram equalization operation. It should be noted that equalization does not necessarily enhance in a favorable direction. The equalization process may also amplify noise interference. Therefore, we use a weighting mechanism to balance information retention and equalization enhancement, where $\alpha = 0.75$. We define the final pixel-level intensity loss function as follows:
\begin{equation}
\mathcal{L}_{\mathrm{int}} = \| I_f - (\alpha \cdot I_{\mathrm{en}} + (1 - \alpha) \cdot I_{\mathrm{eq}} ) \|_1
\end{equation}

\begin{figure}[t]
\centering
\includegraphics[width=1\linewidth]{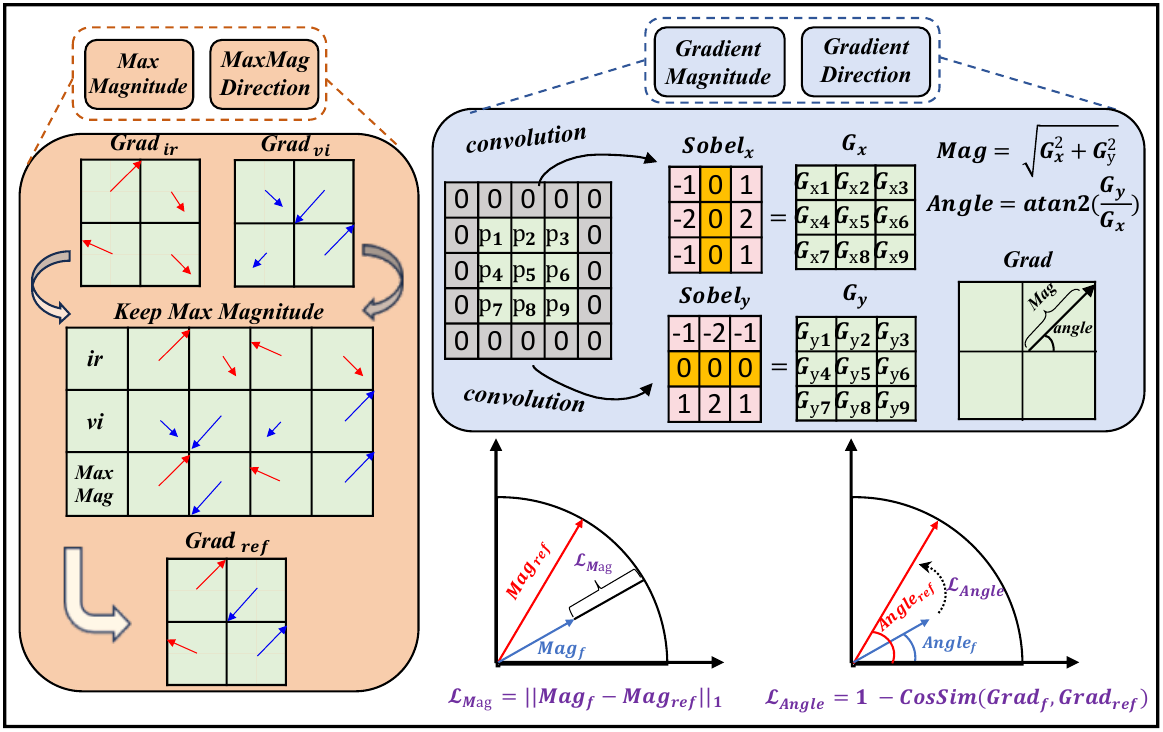}
\caption{The computation flow of the angle-aware loss. First, the Sobel operator is applied to obtain the gradient magnitude and direction of both the fused image and the reference image. Then the maximum-magnitude preservation branch generates the reference gradient $Grad_{ref}$. Finally, a loss that constrains both magnitude and direction is defined to ensure that the fused result maintains better consistency with the source images in texture details.
}
\label{fig1.4}
\vspace{-3mm}
\end{figure}
\subsection{Network Architecture}
As shown in Fig.~\ref{fig1}, we propose a multi-scale fusion network based on U-Net, the network adopts a typical encoder–fusion–decoder structure. We apply the Restormer-CNN block to exploit the advantage of Restormer~\cite{zamir2022restormer} in capturing long-range dependencies while retaining the efficiency of CNN in extracting local features. The network contains four downsampling stages and four upsampling stages. Each stage embeds three Restormer-CNN modules. This enables joint fusion and reconstruction of global context and local details at different scales. The decoder performs progressive upsampling and concatenates with the outputs from the fusion layer at the corresponding scales. 

\subsection{Angle-aware Perception for VIF}
As shown in Fig.~\ref{fig1.4}, we fully consider the vector property of gradients and design loss functions for both direction and magnitude constraints. We first build a gradient vector representation of the image. For each input image $I\in{I_{ir}, I_{vi}, I_{f}}$ the two-dimensional gradient field is extracted using the Sobel operator:
\begin{equation}
\nabla I=[G_x, G_y]\in\mathbb R^{B\times C\times2\times H\times W}
\end{equation}
where $G_x$ and $G_y$ denote the gradient components in the horizontal and vertical directions. Thus the gradients of the infrared image and the visible image are given as follows:
\begin{equation}
\nabla I_{\text{ir}}=[G^{\text{ir}}_x,G^{\text{ir}}_y],
\nabla I_{\text{vi}}=[G^{\text{vi}}_x,G^{\text{vi}}_y]
\end{equation}
The gradient magnitude is defined as the L2 norm of $\nabla I$. Hence the magnitudes of the infrared image and the visible image are expressed as:
\begin{equation}
\begin{aligned}
\mathrm{Mag}_{\mathrm{ir}} &= \| \nabla I_{\mathrm{ir}} \|_2 = \sqrt{\left(G^{\mathrm{ir}}_{x}\right)^2 + \left(G^{\mathrm{ir}}_{y}\right)^2} \\
\mathrm{Mag}_{\mathrm{vi}} &= \| \nabla I_{\mathrm{vi}} \|_2 = \sqrt{\left(G^{\mathrm{vi}}_{x}\right)^2 + \left(G^{\mathrm{vi}}_{y}\right)^2}
\end{aligned}
\end{equation}

\emph{Gradient magnitude loss}: In the image fusion process the gradient reflects pixel intensity variation and is related to key information such as image edges. Regions with larger gradient magnitude contain rich information that is crucial for visual quality. Therefore this information should be fully preserved in both infrared and visible images. We construct a reference gradient $\nabla I_{ref}$ with a max-selection strategy to dynamically select the more important information at each pixel location. The reference gradient $\nabla I_{ref}$ is defined as follows: 
\begin{equation}
\nabla I_{\mathrm{ref}} = 
\begin{cases}
\nabla I_{\mathrm{ir}}, & \text{if } \mathrm{Mag}_{\mathrm{ir}} > \mathrm{Mag}_{\mathrm{vi}} \\
\nabla I_{\mathrm{vi}}, & \text{otherwise}
\end{cases}
\end{equation}
By computation the magnitude of the reference gradient can be obtained as follows:
\begin{equation}
\mathrm{Mag}_{\mathrm{ref}} = \|\nabla I_{\mathrm{ref}}\|_2 = \sqrt{\left(G^{\mathrm{ref}}_x\right)^2 + \left(G^{\mathrm{ref}}_y\right)^2}
\end{equation}
Similarly, the gradient magnitude of the fused image can be computed as follows:
\begin{equation}
\mathrm{Mag}_{f} = \|\nabla I_{f}\|_2 = \sqrt{\left(G^{f}_x\right)^2 + \left(G^{f}_y\right)^2}
\end{equation}
On this basis we define the gradient magnitude loss as follows:
\begin{equation}
\mathcal{L}_{\mathrm{mag}} = \|\mathrm{Mag}_{f} - \mathrm{Mag}_{\mathrm{ref}}\|_2
\end{equation}

\begin{table*}[t]
\small
\centering
 \caption{Test results on the MSRS dataset. The optimal, second-optimal, and third-optimal values are labelled in red, blue, and green respectively.}
\begin{tabular}{lllllllll}
\hline
                  & EN $\uparrow$      & SD$\uparrow$    & SF$\uparrow$   & AG$\uparrow$    & SCD$\uparrow$ & VIF$\uparrow$& Qabf$\uparrow$ & SSIM$\uparrow$ \\ \hline
CDDFuse~\cite{Zhao_2023_CVPR}     & 6.685    & 42.986   & 11.729   & 3.804    & 1.602 & \textcolor{red}{1.053} & \textcolor{red}{0.719}  & 0.694  \\
Diff-IF~\cite{diff}     & 6.669    & 42.598   & 11.459   & 3.696    & 1.624 & \textcolor{green}{1.042} & 0.685  & \textcolor{red}{0.699}  \\
AGMFusion~\cite{agm}   & \textcolor{blue}{6.758}    & \textcolor{green}{44.503}   & \textcolor{blue}{11.712}   & \textcolor{blue}{4.172}    & \textcolor{red}{1.814} & 0.881 & 0.574  & 0.683  \\
EMMA~\cite{Zhao_2024_CVPR}        & 6.718    & \textcolor{blue}{44.577}   & 11.554   & 3.779    & 1.629 & 0.973 & 0.642  &\textcolor{red}{ 0.699 } \\
SMAE-Fusion~\cite{smae} & \textcolor{green}{6.719}    & 43.760    & \textcolor{green}{11.696}   & \textcolor{green}{3.840}     & \textcolor{blue}{1.701} & 1.080  & 0.687  & 0.683  \\
Conti-Fuse~\cite{conti}  & 6.654    & 42.705   & 11.512   & 3.690     & 1.639 & \textcolor{blue}{1.044} & \textcolor{blue}{0.706}  & \textcolor{blue}{0.698}  \\
KDfuse~\cite{kdfuse}      & 6.655    & 42.060    & 10.984   & 3.676    & 1.617 & 1.039 &\textcolor{green}{ 0.705}  & \textcolor{green}{0.695}  \\ \hline \hline
\bfseries Ours              & \bfseries\textcolor{red}{7.122(5\%$\uparrow$)} &\bfseries\textcolor{red}{ 47.429(6\%$\uparrow$)} & \bfseries\textcolor{red}{12.596(7\%$\uparrow$)} & \bfseries\textcolor{red}{4.300(3\%$\uparrow$)} & \bfseries\textcolor{green}{1.641} &\bfseries 1.010  &\bfseries 0.646  &\bfseries 0.633  \\
\hline
\end{tabular}
 \label{tab1}
\end{table*}

\begin{table*}[t]
\small 
\centering
 \caption{Test results on the M3FD dataset. The optimal, second-optimal, and third-optimal values are labelled in red, blue, and green respectively.}
\begin{tabular}{lllllllll}
\hline
                  & EN$\uparrow$ & SD$\uparrow$ & SF$\uparrow$ & AG$\uparrow$ & SCD$\uparrow$ & VIF$\uparrow$ & Qabf$\uparrow$ & SSIM$\uparrow$ \\ \hline
CDDFuse~\cite{Zhao_2023_CVPR}     & 7.083        & 41.220        & \textcolor{blue}{16.825}       & 5.552        & 1.507         &\textcolor{green}{ 0.808}         & \textcolor{green}{0.655}          & \textcolor{blue}{0.692}          \\
Diff-IF~\cite{diff}    & 6.987        & 38.747       & 15.833       & 5.143        & 1.354         & 0.766         & 0.598          & 0.684          \\
AGMFusion~\cite{agm}   & \textcolor{green}{7.127}        & 41.482       & 15.611       & \textcolor{green}{5.601}        & \textcolor{red}{1.789}         & 0.668         & 0.536          & 0.646          \\
EMMA~\cite{Zhao_2024_CVPR}        & 7.123        & \textcolor{green}{42.993}       & \textcolor{green}{16.778}       & \textcolor{blue}{5.856}        & 1.523         & 0.760          & 0.614          & 0.689          \\
SMAE-Fusion~\cite{smae} & \textcolor{blue}{7.180}         & \textcolor{blue}{43.947}       & 16.074       & 5.349        & \textcolor{green}{1.640}          & 0.781         & 0.624          & \textcolor{red}{0.697}          \\
Conti-Fuse~\cite{conti}  & 6.977        & 38.414       & 16.045       & 5.097        & 1.396         & 0.768         & 0.617          & \textcolor{green}{0.690}           \\
KDfuse~\cite{kdfuse}     & 7.010         & 38.904       & 15.183       & 5.311        & 1.368         &\textcolor{blue}{ 0.822  }       &\textcolor{red}{ 0.673  }        & 0.689          \\ \hline \hline
\bfseries Ours              &\bfseries\textcolor{red}{ 7.286(1\%$\uparrow$)}        & \bfseries\textcolor{red}{44.091(0.3\%$\uparrow$)}       & \bfseries\textcolor{red}{17.063(1\%$\uparrow$)}       & \bfseries\textcolor{red}{6.052(3\%$\uparrow$)}        & \bfseries\textcolor{blue}{1.681 }        & \bfseries\textcolor{red}{0.844(2\%$\uparrow$) }        & \bfseries\textcolor{blue}{0.661}          &\bfseries\textcolor{green}{ 0.690}           \\ \hline
\end{tabular}
 \label{tab2}
\end{table*}

\begin{table*}[t]
\small
\centering
 \caption{Test results on the RoadScene dataset. The optimal, second-optimal, and third-optimal values are labelled in red, blue, and green respectively.}
\begin{tabular}{lllllllll}
\hline
                  & EN$\uparrow$ & SD$\uparrow$ & SF$\uparrow$ & AG$\uparrow$ & SCD$\uparrow$ & VIF$\uparrow$ & Qabf$\uparrow$ & SSIM$\uparrow$ \\ \hline
CDDFuse~\cite{Zhao_2023_CVPR}     & 7.432        & 50.165       & \textcolor{blue}{17.125}       & \textcolor{blue}{6.268}        & 1.533         & 0.636         & \textcolor{green}{0.559 }         & 0.674          \\
Diff-IF
~\cite{diff}     & 7.210         & 44.594       & 12.998       & 4.875        & 1.223         & \textcolor{green}{0.669}         & 0.512          & 0.675          \\
AGMFusion~\cite{agm}   & 7.204        & 50.507       & 13.962       & 5.478        & 1.530          & 0.543         & 0.414          & 0.634          \\
EMMA~\cite{Zhao_2024_CVPR}        & \textcolor{blue}{7.515}        & \textcolor{blue}{54.402}       & 15.079       & 5.741        & \textcolor{blue}{1.676}         & 0.642         & 0.462          & 0.666          \\
SMAE-Fusion~\cite{smae} & \textcolor{green}{7.504}        & \textcolor{red}{54.605}       & \textcolor{green}{16.347}       & \textcolor{green}{6.054}        & \textcolor{red}{1.686}         & \textcolor{blue}{0.683}         & \textcolor{blue}{0.560}           & \textcolor{red}{0.684}          \\
Conti-Fuse~\cite{conti}  & 7.210         & 43.888       & 15.747       & 5.559        & 1.247         & 0.658         & 0.531          & \textcolor{green}{0.677}          \\
KDfuse~\cite{kdfuse}      & 7.302        & 45.596       & 13.830        & 5.690         & 1.287         & 0.653         & 0.549          & \textcolor{blue}{0.678}          \\ \hline \hline
\bfseries Ours              & \bfseries\textcolor{red}{7.545(0.3\%$\uparrow$)}        & \bfseries\textcolor{green}{51.671}      & \bfseries\textcolor{red}{18.327(7\%$\uparrow$)}       & \bfseries\textcolor{red}{7.000(11\%$\uparrow$)}            & \bfseries\textcolor{green}{1.614}         & \bfseries\textcolor{red}{0.707(3\%$\uparrow$)}         & \bfseries\textcolor{red}{0.618(10\%$\uparrow$)}          & \bfseries0.675 \\
\hline
\end{tabular}
 \label{tab3}
\end{table*}

\emph{Gradient direction loss}: To further constrain the gradient orientation, we introduce a direction constraint based on cosine similarity. From the perspective of vector space, the angular deviation between two gradient fields can be represented by the dot product, as follows:
\begin{equation}
\mathrm{CosSim}(\nabla I_{\mathrm{ref}}, \nabla I_f) =
\frac{\langle \nabla I_{\mathrm{ref}}, \nabla I_f \rangle}
{\|\nabla I_{\mathrm{ref}}\|_2 \, \|\nabla I_f\|_2}
\end{equation}
where $\langle \cdot, \cdot \rangle$ represents the inner product. When the angular deviation between two vectors is smaller, the value of $Cos\_Sim(\cdot,\cdot)$ becomes larger, with a range of [-1,1]. Thus the gradient direction loss is defined as follows:
\begin{equation}
\mathcal L_{\text{angle}}=1- {CosSim}(\nabla I_{ref}, \nabla I_f)
\end{equation}

Finally, the complete gradient loss function is defined as a linear combination of the magnitude loss and the direction loss:
\begin{equation}
{\mathcal L_{\text{grad}}=\lambda_{1} \cdot \mathcal L_{\text{mag}}+\lambda_{2} \cdot \mathcal L_{\text{angle}}}
\end{equation}
where $\lambda_{1}$ and $\lambda_{2}$ are a set of hyperparameters with values $\lambda_{1}=5$ and $\lambda_{2}=1$. The overall implementation is shown in Algorithm.~\ref{alg1}. Therefore, our total loss is defined as follows:
\begin{equation}
{\mathcal L_{\text{total}}=\mathcal L_{\text{int}}+\ \mathcal L_{\text{grad}}}
\end{equation}

\begin{figure*}[!ht]
\centering
\includegraphics[width=0.95\linewidth]{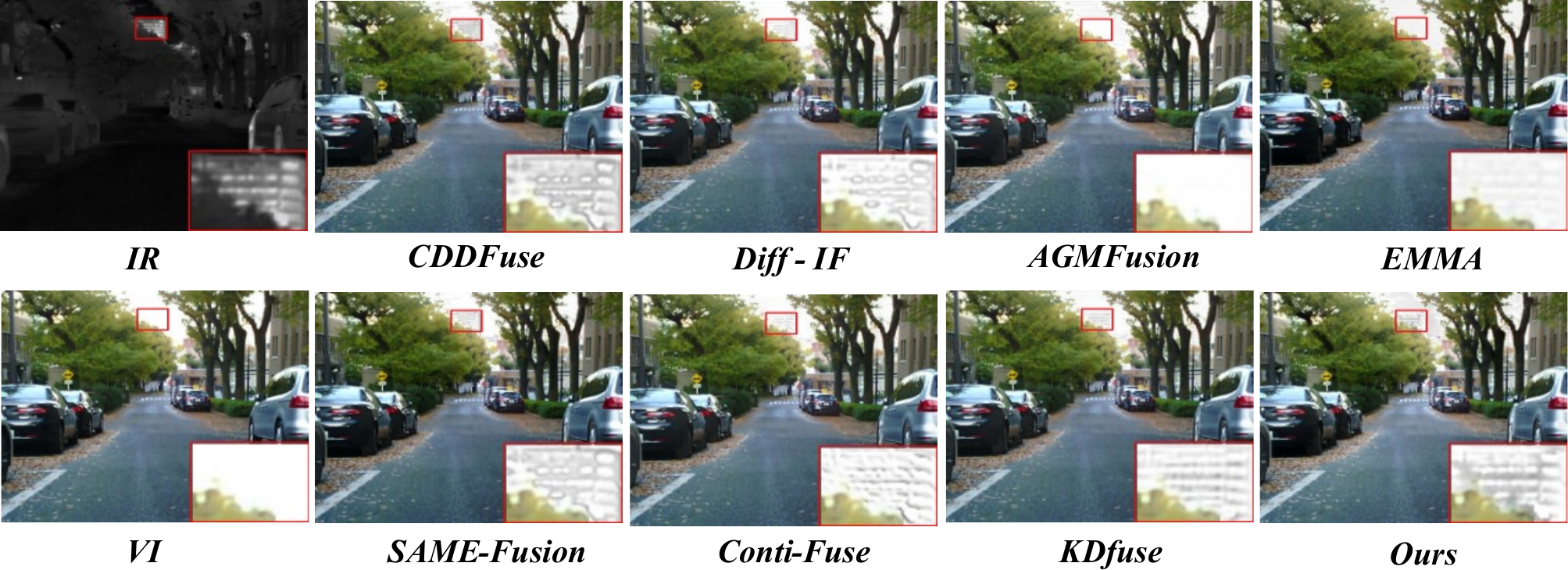}
\caption{Fusion visualization results on the MSRS test set}
\label{fig2}
\vspace{-3mm}
\end{figure*}

\begin{figure*}[!t]
\centering
\includegraphics[width=0.95\linewidth]{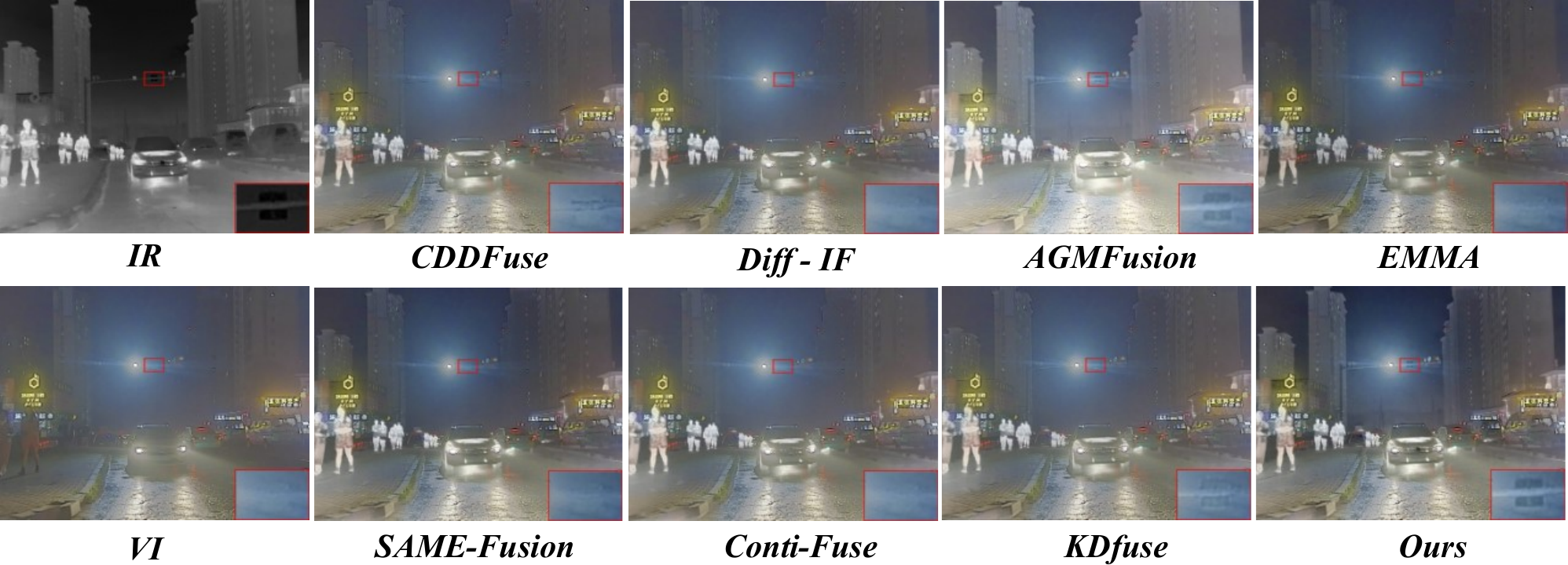}
\caption{Fusion visualization results on the M3FD test set. In complex urban scenes with strong light sources, our method highlights thermal targets while preserving texture details.}
\label{fig3}
\vspace{-3mm}
\end{figure*}

\begin{figure*}[!t]
\centering
\includegraphics[width=0.95\linewidth]{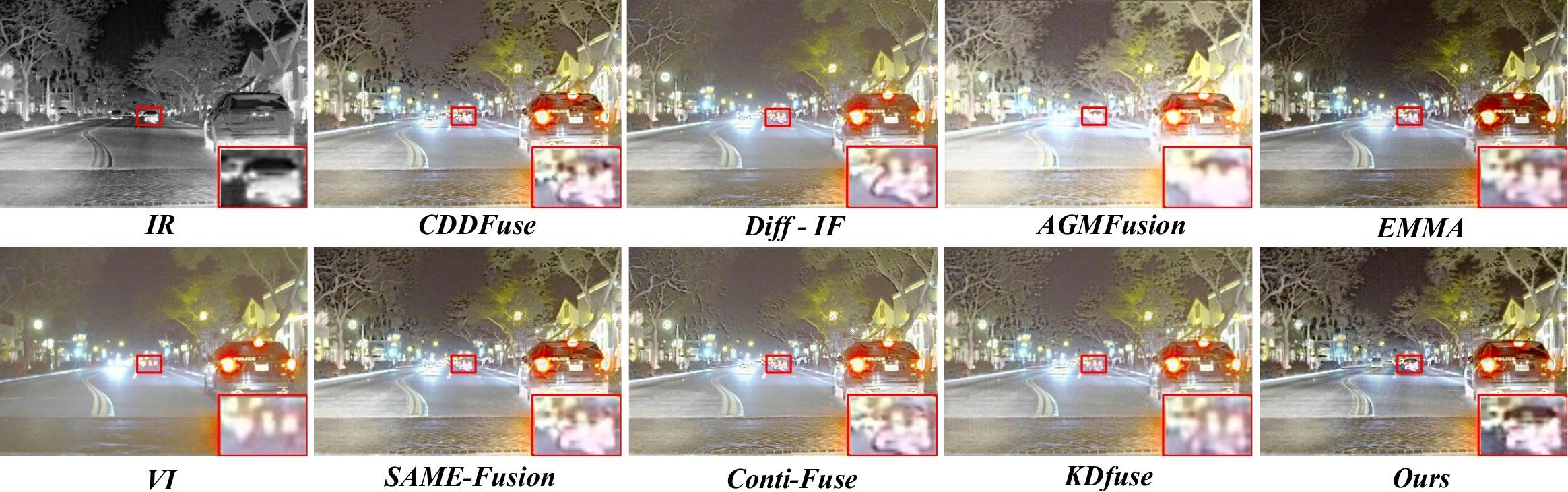}
\caption{Fusion visualization results on the RoadSence test set. Our method yields sharper edges and complete detail with uniform natural brightness.}
\label{fig4}
\vspace{-3mm}
\end{figure*}

\begin{table*}[t]
\normalsize
\centering
 \caption{Ablation study results. Red indicates the best value and blue indicates the second-best value.}
\begin{tabular}{c|ccc|cccccccc}
\hline
\multirow{2}{*}{datasets}  & \multirow{2}{*}{ComMask} & \multirow{2}{*}{AngularLoss} & \multirow{2}{*}{FRIS} & \multicolumn{8}{c}{Performance}                                                                                             \\ \cline{5-12} 
                           &                          &                              &                       & EN$\uparrow$ & SD$\uparrow$ & SF$\uparrow$ & AG$\uparrow$ & SCD$\uparrow$ & VIF$\uparrow$ & Qabf$\uparrow$ & SSIM$\uparrow$ \\ \hline 
\multirow{4}{*}{MSRS}      & $\times$                 & $\times$                     & $\times$              & 6.673        & 42.918       & 11.551       & 3.749        & 1.608         & 1.062         & 0.735          & 0.694          \\
                           & $\checkmark$             & $\times$                     & $\times$              & 6.675        & 42.922       & 11.565       & 3.750         & 1.611         & 1.065         & 0.736          & 0.695          \\
                           & $\checkmark$             & $\checkmark$                 & $\times$              & 6.698        & 42.702       & 11.639       & 3.833        & 1.737         & 0.957         & 0.662          & 0.706          \\
                           & $\checkmark$             & $\checkmark$                 & $\checkmark$          &\bfseries\textcolor{red}{7.122}        & \bfseries\textcolor{red}{47.429}       & \bfseries\textcolor{red}{12.596}       & \bfseries\textcolor{red}{4.300}          & \bfseries\textcolor{blue}{1.641   }      & \bfseries 1.010          &\bfseries 0.646          &\bfseries 0.633          \\ \hline
\multirow{4}{*}{RoadSence} & $\times$                 & $\times$                     & $\times$              & 7.411        & 49.470        & 16.875       & 6.287        & 1.311         & 0.779         & 0.641          & 0.653          \\
                           & $\checkmark$             & $\times$                     & $\times$              & 7.454        & 50.768       & 17.082       & 6.421        & 1.339         & 0.772         & 0.641          & 0.648          \\
                           & $\checkmark$             & $\checkmark$                 & $\times$              & 7.475        & 49.391       & 17.608       & 6.670         & 1.522         & 0.701         & 0.620           & 0.687          \\
                           & $\checkmark$             & $\checkmark$                 & $\checkmark$          & \bfseries\textcolor{red}{7.545}        & \bfseries\textcolor{red}{ 51.671}      & \bfseries\textcolor{red}{18.327}      &\bfseries\textcolor{red}{ 7.000}            &\bfseries\textcolor{red}{ 1.614}        &\bfseries 0.707         & \bfseries0.618          &\bfseries\textcolor{blue}{ 0.670  }        \\ \hline
\multirow{4}{*}{M3FD}      & $\times$                 & $\times$                     & $\times$              & 7.047        & 40.214       & 16.470        & 5.512        & 1.317         & 0.88          & 0.696          & 0.682          \\
                           & $\checkmark$             & $\times$                     & $\times$              & 7.052        & 40.362       & 16.572       & 5.477        & 1.338         & 0.865         & 0.688          & 0.678          \\
                           & $\checkmark$             & $\checkmark$                 & $\times$              & 7.095        & 39.595       & 16.461       & 5.627        & 1.580          & 0.817         & 0.667          & 0.700            \\
                           & $\checkmark$             & $\checkmark$                 & $\checkmark$          & \bfseries\textcolor{red}{7.286 }       & \bfseries\textcolor{red}{44.091}       & \bfseries\textcolor{red}{17.063 }      & \bfseries\textcolor{red}{6.052}     & \bfseries\textcolor{red}{1.681}       & \bfseries 0.844         &\bfseries 0.661          &\bfseries\textcolor{blue}{ 0.690}         \\ \hline
                 
\end{tabular}
 \label{tab5}
\end{table*}

\begin{table}[t]
\normalsize
\centering
 \caption{Efficiency analysis of different image fusion methods.}
\begin{tabular}{lll}
\hline
Method      & Params (M) & Inference times (ms) \\\hline
CDDFuse~\cite{Zhao_2023_CVPR}     & 1.188      & 236.77               \\
Diff-IF~\cite{diff}     & 23.736     & 1826.60               \\
AGMFusion~\cite{agm}   & 13.841     & 4.62                \\
EMMA~\cite{Zhao_2024_CVPR}        & 1.518      & 34.67                \\
SMAE-Fusion~\cite{smae} & 1.921      & 673.92               \\
Conti-Fuse~\cite{conti}  & 1.661      & 55.67               \\
KDfuse~\cite{kdfuse}      & 4.000          & 211.08               \\      \hline\hline
Ours        & 1.623      & 26.42      \\
\hline         
\end{tabular}
 \label{tab6}
\end{table}

\begin{figure}[!t]
\centering
\includegraphics[width=0.95\linewidth]{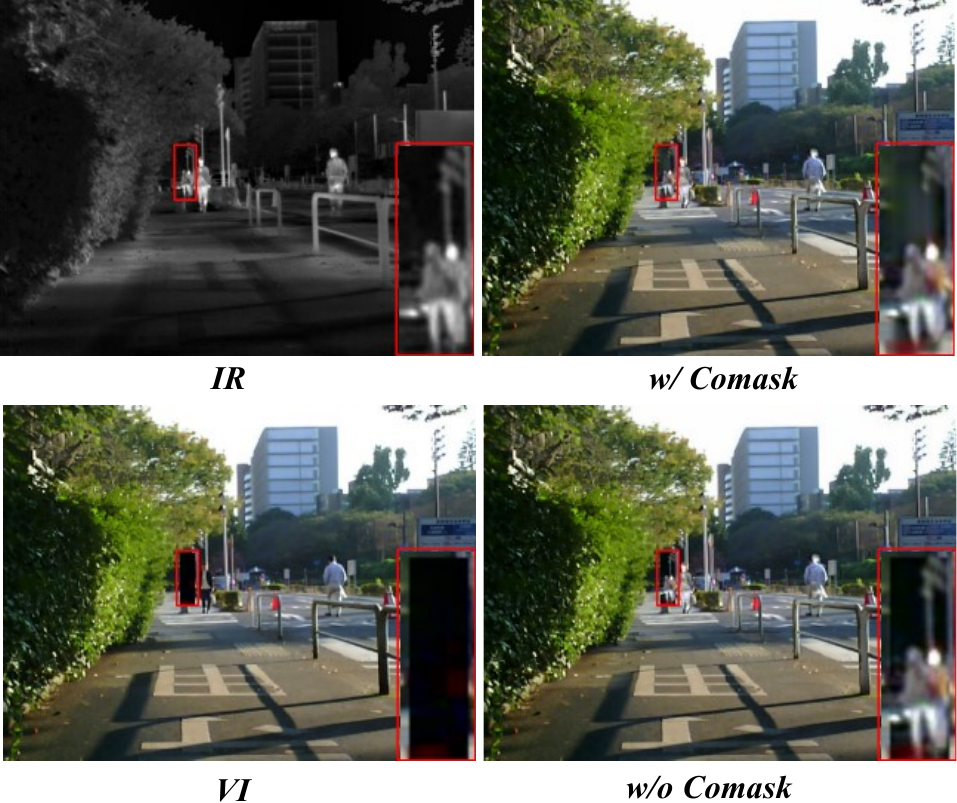}
\caption{The test results under simulated information-missing scenarios. With the introduction of ComMask the network exploits cross-modal complementary information and maximizes information recovery from the other modality under missing information conditions.}
\label{fig5}
\vspace{-3mm}
\end{figure}

\begin{figure}[!t]
\centering
\includegraphics[width=0.95\linewidth]{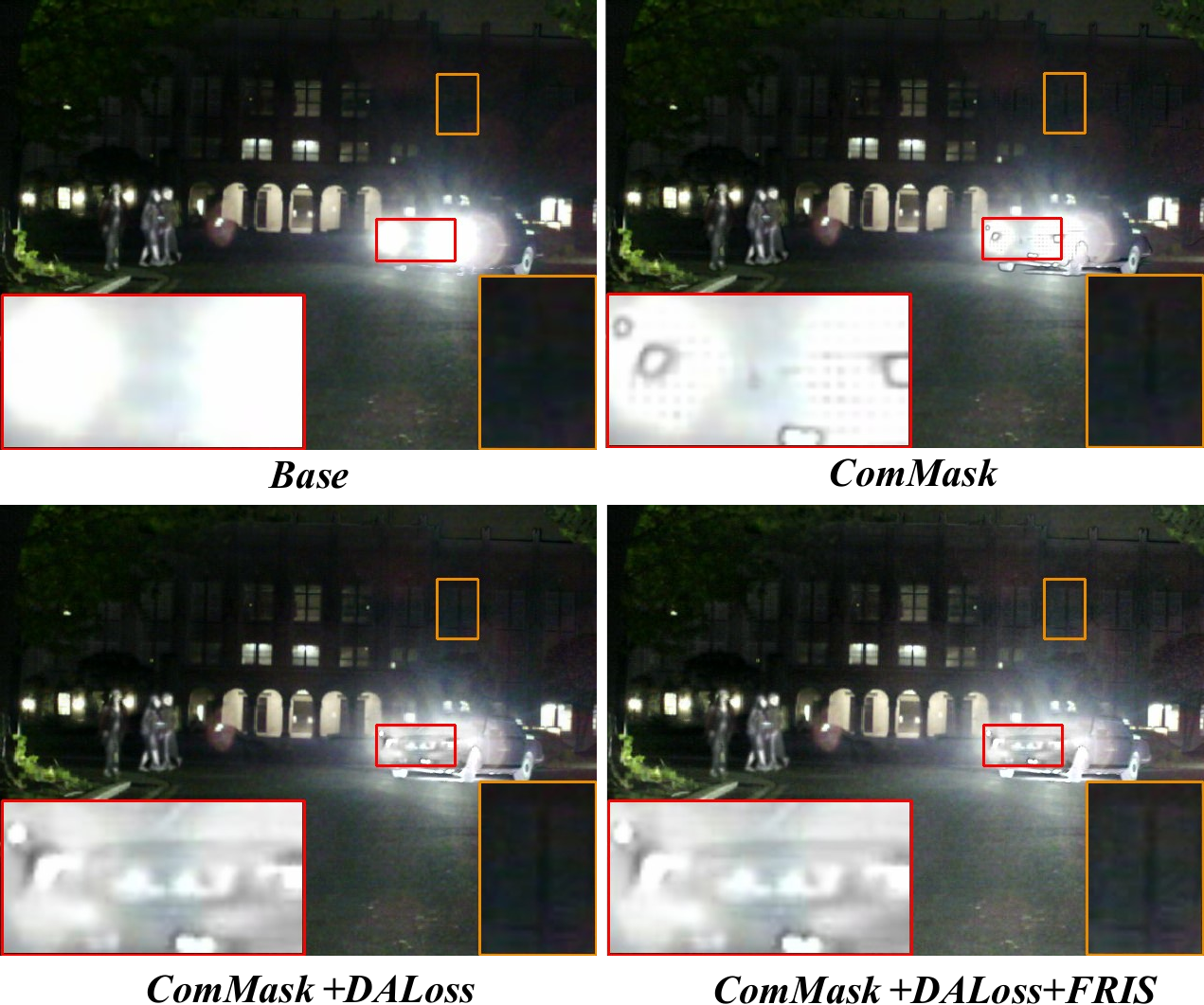}
\caption{Visualization results of the ablation study. With the gradual addition of the proposed modules the vehicle contours in the red region become clearer. The brightness distribution in the orange box region becomes more uniform which makes the visual perception more natural.}
\label{fig6}
\vspace{-3mm}
\end{figure}

\begin{algorithm}
\caption{Angle-aware Perceptual Loss Computation}
\begin{algorithmic}
\State \textbf{Input:}  Infrared image $I_{ir}$, Visible image $I_{vi}$ and Fused image $I_f$
\State \textbf{Output:}  Angle-aware Perceptual Loss $L_{grad}$

\State $\triangleright$ Compute gradient fields:
    \For{each image $I \in \{I_{ir}, I_{vi}, I_f\}$}
  \State  $G_x=Sobel_{x}(I),\quad G_y=Sobel_{y}(I)$ 

   \State Form gradient field $\nabla I = [G_x, G_y] \in \mathbb{R}^{B \times C \times 2 \times H \times W}$
    \EndFor

\State $\triangleright$ Compute gradient magnitudes:
\For{each image $I \in \{I_{ir}, I_{vi}\}$}
    \State $Mag_{ir} = \|\nabla I_{ir}\|_2 = \sqrt{(G^{\text{ir}}_x)^2 + (G^{\text{ir}}_y)^2}$
    \State $Mag_{vi} = \|\nabla I_{vi}\|_2 = \sqrt{(G^{\text{vi}}_x)^2 + (G^{\text{vi}}_y)^2}$
\EndFor

\State$\triangleright$ Construct reference gradient:
\For{each pixel location}
    \If{$Mag_{ir} > Mag_{vi}$}
        \State $\nabla I_{ref} = \nabla I_{ir}$
    \Else
        \State $\nabla I_{ref} = \nabla I_{vi}$
    \EndIf
\EndFor
\State$\triangleright$ Compute reference gradient magnitude: 
\State$Mag_{ref} = \|\nabla I_{ref}\|_2 = \sqrt{(G^{ref}_x)^2 + (G^{ref}_y)^2}$

\State$\triangleright$ Compute fused image gradient magnitude:
\State $Mag_f = \|\nabla I_f\|_2 = \sqrt{(G^f_x)^2 + (G^f_y)^2}$

\State$\triangleright$ Compute Angle-aware Perceptual loss:  
\State $L_{mag} = \| Mag_f - Mag_{ref} \|_2$

\State $CosSim(\nabla I_{ref}, \nabla I_f) = \frac{\langle \nabla I_{ref}, \nabla I_f \rangle}{\|\nabla I_{ref}\|_2 \|\nabla I_f\|_2}$
\State $L_{angle} = 1 - CosSim(\nabla I_{ref}, \nabla I_f)$

\State $L_{grad} = \lambda_1 \cdot L_{mag} + \lambda_2 \cdot L_{angle}$

\State \Return $L_{grad}$
\end{algorithmic}
\label{alg1}
\end{algorithm}

\section{Experimental Validation}

\subsection{Setup}
Our method is primarily implemented using PyTorch and experiments are conducted on an NVIDIA GeForce RTX 3090. We evaluate our approach on three representative public datasets: RoadScene, MSRS, and M3FD. During training we use 1083 image pairs from the MSRS training set. All training samples are randomly cropped into patches of size 128×128 with a batch size of 4, and the model is trained for 70 epochs. We employ the Adam optimizer with an initial learning rate of 1e-4. A cosine annealing schedule is used to adjust the learning rate, with a minimum value of 1e-6.

Out-of-Distribution Evaluation: To validate fusion performance and generalization, we tested the model on 70 RoadScene pairs and 100 M3FD pairs that were not used in training.

\subsection{Comparisons}
On the MSRS test set, we compared our method with seven representative SOTA algorithms published between 2023 and 2025. These methods include CDDfuse~\cite{Zhao_2023_CVPR}, Diff-IF~\cite{diff},, AGMfusion~\cite{agm}, EMMA~\cite{Zhao_2024_CVPR}, SMAE-Fusion~\cite{smae}, Conti-Fuse~\cite{conti}, and KDfuse~\cite{kdfuse}.

Table.~\ref{tab1} indicates that our method surpasses the second-rank approach by 5\% in EN, 6\% in SD, 7\% in SF, and 3\% in AG. The gains confirm richer content and enhanced texture detail in the fused images. From the visualization results in Fig.~\ref{fig2} it can be observed that our method shows significant advantages in the image fusion task. In the regions marked by red boxes the results demonstrate better preservation of texture details. The fusion results of our method present richer details of the buildings.

\subsection{Out-of-Distribution Experiment}
As shown in Table.~\ref{tab2} and Table.~\ref{tab3} our method still achieves state-of-the-art fusion results on the RoadScene and M3FD datasets that are not used for training. These results clearly demonstrate that the proposed method can still generate fused images with richer information even in Out-of-Distribution (O.O.D) scenarios, showing strong cross-domain generalization ability.

Fig.~\ref{fig3} shows the image fusion results on the M3FD dataset. In the regions marked with red boxes it can be observed that our method achieves better balance of contrast and brightness. The contours and details of the targets in the visible image are enhanced. Other methods show clear performance degradation when handling dark backgrounds and fail to effectively separate the targets from the background.

As shown in Fig.~\ref{fig4} other methods lose target details to varying degrees due to illumination issues. In contrast our method clearly depicts the contours of vehicles and achieves high-quality fusion in many details of the image.

\subsection{Ablations}
In this ablation study we systematically evaluate the impact of three modules, ComMask, AngularLoss (Angle-aware perceptual loss), and FRIS (Fine-Grained Reference Image Synthesis), on image fusion performance. Experiments are conducted on the MSRS, RoadScene, and M3FD datasets. Experimental results show that progressively introducing ComMask, AngularLoss, and FRIS leads to improvements across different objective metrics. The combined effect of these improvements enables the fused images to outperform the baseline methods across multiple evaluation metrics. This validates the effectiveness of each module in enhancing image fusion quality.

\textbf{Ablation on ComMask.} The experimental results in Table.~\ref{tab5} show that the ComMask module can effectively integrate multi-modal information and improve image quality in the image fusion task. On the MSRS dataset all objective evaluation metrics increase. This demonstrates its effectiveness and superiority in image fusion. Metrics on the OOD test sets RoadScene and M3FD also show stable improvement. This indicates that the module has strong generalization performance. In the regions marked by red boxes in Fig.~\ref{fig5}, the method with ComMask not only utilizes the information within the image more effectively but also successfully integrates information from the other modality.

\textbf{Ablation on Angle-aware Perception.} The experimental results in Table.~\ref{tab5} show that adding AngularLoss improves the performance of image fusion metrics. Several objective indicators including EN, SF, AG, SCD, and SSIM are enhanced. This indicates that AngularLoss through joint constraints on magnitude and direction effectively guides the fused image to retain richer gradient information from the source images. As shown in the regions marked by red boxes in Fig.~\ref{fig6}, introducing AngularLoss significantly improves target details and makes the contours of vehicles clearer. This shows that AngularLoss can effectively promote the algorithm to capture and preserve details during fusion and improve the visual quality and information content of the final image.

\textbf{Ablation on Fine-Grained Reference Image Synthesis (FRIS).} The experimental results in Table.~\ref{tab5} show that further introducing FRIS leads to more significant improvement in model performance. The EN, SD, SF, and AG metrics all achieve SOTA performance, which fully validates the effectiveness of the FRIS module. The effect is shown in the regions marked by orange boxes in Fig.~\ref{fig6}. By extracting high-frequency texture information with the Laplacian operator and combining it with histogram equalization, FRIS effectively mitigates uneven brightness.

\textbf{Efficiency analysis:} We conduct efficiency analysis of several image fusion methods with a focus on model parameters and inference time as the two key indicators. As shown in Table.~\ref{tab6}, although CDDFuse has the smallest number of parameters (1.188M), its inference time reaches 236.77 ms. This may limit its practicality in applications requiring fast response. Our model achieves a good trade-off between parameter count and inference time. It has an inference time of 26.42 ms while maintaining a low parameter count of 1.623M. This achieves a good trade-off between performance and efficiency. The experimental results further demonstrate that the proposed model not only achieves higher accuracy but also offers strong practicality.

\section{Conclusion}
This work addresses the limitations of intensity loss and gradient loss in image fusion and proposes the AngularFuse framework. First, a complementary mask learning mechanism is used to encourage the network to explore cross-modal complementary information. Second, a fine-grained reference image synthesis module is designed. It extracts high-frequency information using the Laplacian operator and combines it with histogram equalization to dynamically generate scene-adaptive target images. Finally, the vector property of gradients is fully considered. We propose a angle-constrained gradient loss that enforces consistency in both magnitude and direction. Comprehensive experiments on the MSRS, RoadScene, and M3FD datasets show that AngularFuse achieves superior performance across multiple objective metrics.

\bibliographystyle{IEEEtran}
\bibliography{ref}

\begin{IEEEbiography}
[{\includegraphics[width=1in,height=1.25in,clip,keepaspectratio]{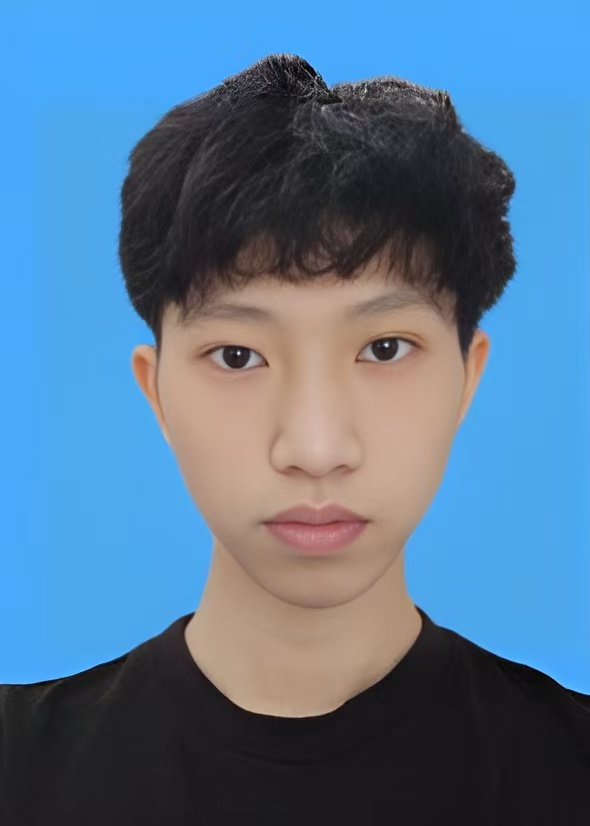}}]{Xiaopeng Liu } received the B.S. degree in 2024, from the School of Information Engineering, Guangdong University of Technology, Guangzhou, China, where he is currently working towards a M.S. degree. His research interests include computer vision and machine learning
\end{IEEEbiography}

\begin{IEEEbiography}[{\includegraphics[width=1in,height=1.25in,clip,keepaspectratio]{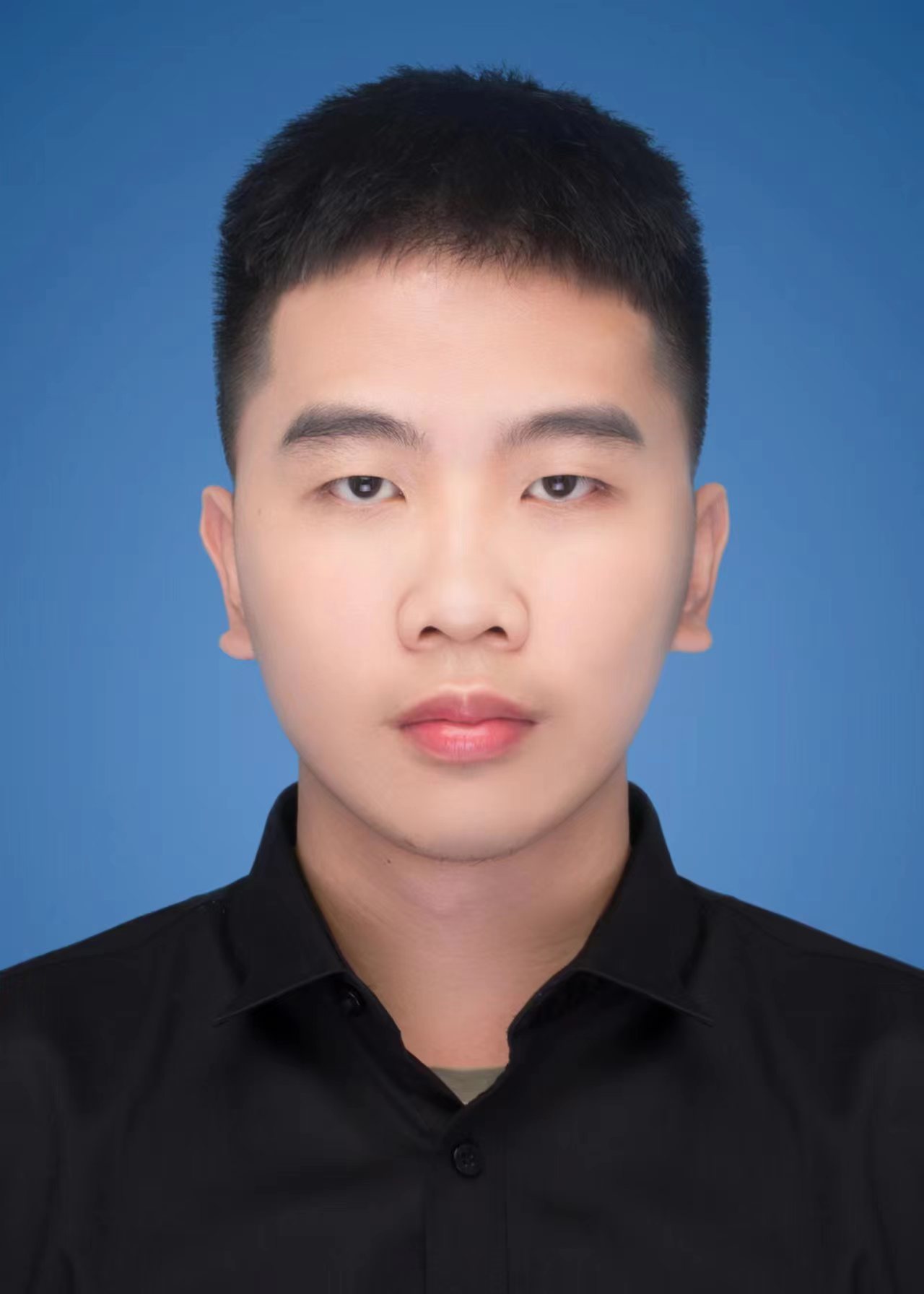}}]{Yupei Lin } received the M.S. degree in 2025, from the School of Information Engineering, Guangdong University of Technology, Guangzhou, China. Currently, he is working towards a Ph.D degree at the School of Computer Science and Engineering, Sun Yat-sen University, Guangzhou, China. His research interests include computer vision and machine learning
\end{IEEEbiography}

\begin{IEEEbiography}
[{\includegraphics[width=1in,height=1.25in,clip,keepaspectratio]{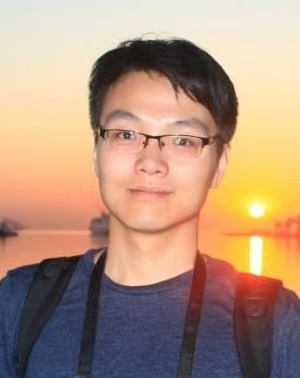}}]{Sen Zhang}
received the Ph.D. degree from the School of Computer Science, the University of Sydney, Australia, in 2023. Previously, he obtained a B.S. degree in 2014, from the School of Biomedical Engineering, Tsinghua University, China. He is currently a Researcher at TikTok, ByteDance Inc, Australia. His research interests include reinforcement learning, computer vision and large language model.
\end{IEEEbiography}

\begin{IEEEbiography}
[{\includegraphics[width=1in,height=1.25in,clip,keepaspectratio]{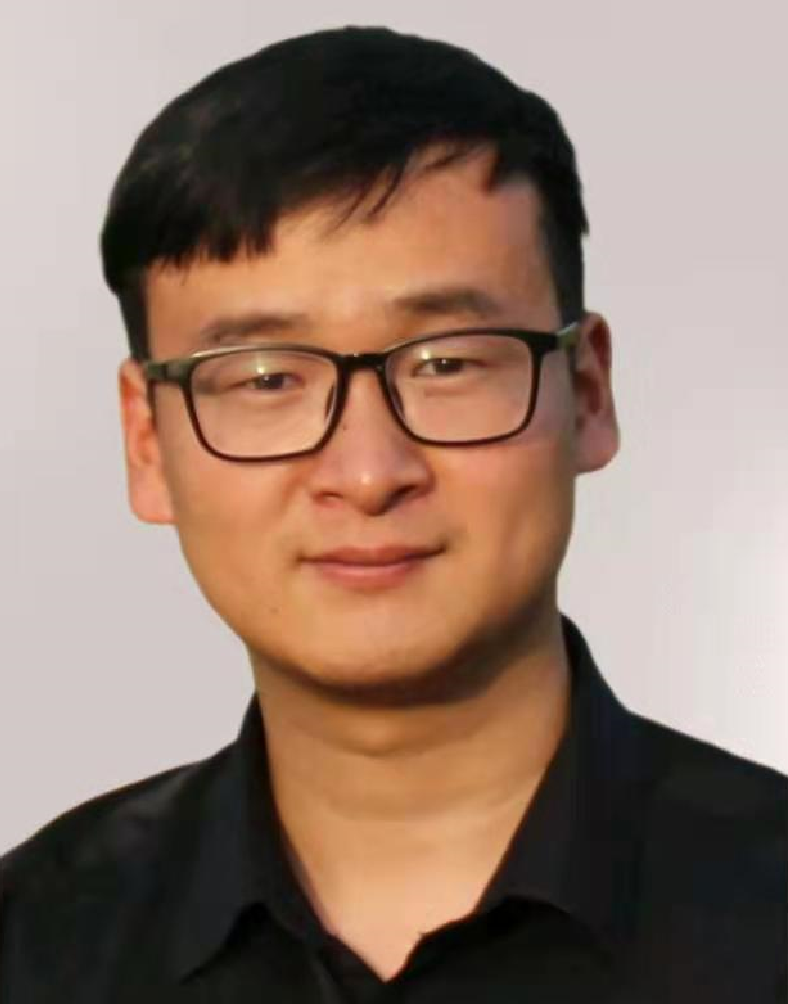}}]{Xiao Wang}
received the PhD degree in computer science from Anhui University, Hefei, China, in 2019. He finished the postdoc research with Peng Cheng Laboratory, Shenzhen, China, from April 2020 to April 2022. He is now an associate professor with the School of Computer Science and Technology, Anhui University, Hefei, China. His current research interests mainly include computer vision, event-based vision, machine learning, and pattern recognition. 
\end{IEEEbiography}

\begin{IEEEbiography}
[{\includegraphics[width=1in,height=1.25in,clip,keepaspectratio]{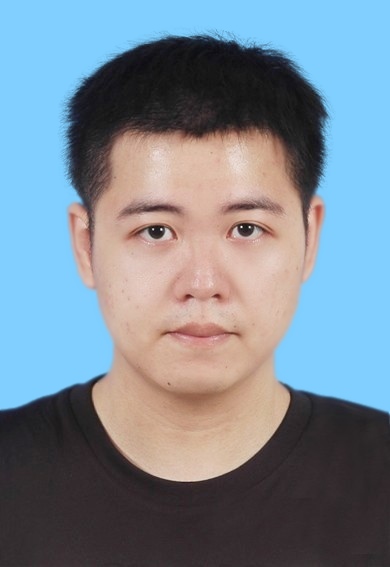}}]{Yukai Shi}
received the Ph.D. degree from the School of Data and Computer Science, Sun Yat-sen University, Guangzhou China, in 2019. He is currently an associate professor at the School of Information Engineering, Guangdong University of Technology, China. His research interests include computer vision and machine learning.
\end{IEEEbiography}

\begin{IEEEbiography}
[{\includegraphics[width=1in,height=1.25in,clip,keepaspectratio]{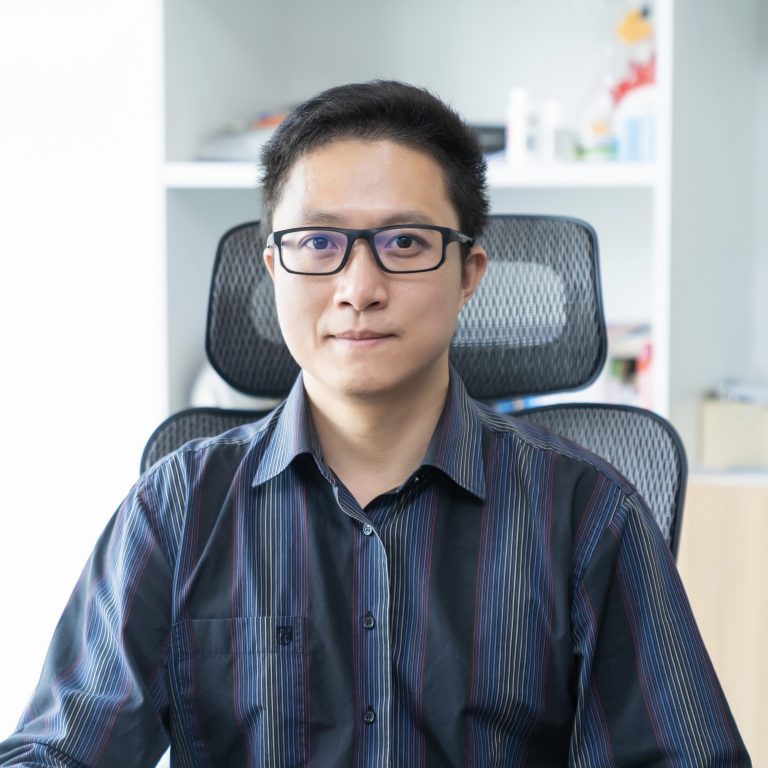}}]{Liang Lin}(Fellow, IEEE) is a Full Professor of computer science at Sun Yat-sen University. He is an associate editor of IEEE T-NNLS and IEEE T-MM, and served as Area Chairs for numerous conferences such as CVPR, ICCV, SIGKDD and AAAI. He is the recipient of numerous awards and honors including Wu Wen-Jun Artificial Intelligence Award, ICCV Best Paper Nomination in 2019, Annual Best Paper Award by Pattern Recognition (Elsevier) in 2018, Google Faculty Award in 2012.
\end{IEEEbiography}

\end{document}